%% file: main.tex
\documentclass[runningheads]{llncs}
\usepackage[T1]{fontenc}
\usepackage{graphicx}
\usepackage{longtable}
\usepackage{multirow}
\usepackage{multicol}
\usepackage{tabularx}
\usepackage[export]{adjustbox}
\usepackage{subcaption}
\usepackage{amsmath}
\usepackage{multibib}
\usepackage{moresize}
\usepackage{graphicx}
\usepackage[usenames,dvipsnames]{color}

\begin{document}
\title{Machine Learning for Raman Spectroscopy-based Cyber-Marine Fish Biochemical Composition Analysis}
\titlerunning{Machine Learning for Raman Spectroscopy in Fish Composition Analysis}
%
\author{
Yun Zhou\inst{1} \and
Gang Chen\inst{1} \and
Bing Xue\inst{1} \and
Mengjie Zhang\inst{1} \and
Jeremy S. Rooney\inst{2} \and
Kirill Lagutin\inst{3} \and
Andrew MacKenzie\inst{3} \and
Keith C. Gordon\inst{3} \and
Daniel P. Killeen\inst{4} 
}
\authorrunning{Yun et al.}
%
\institute{Centre for Data Science and Artificial Intelligence \& School of ECS, Victoria University of Wellington, New Zealand \\
\email{\{zhouyun, aaron.chen, bing.xue, mengjie.zhang\}@ecs.vuw.ac.nz}\\
\and
Department of Chemistry, University of Otago, Dunedin, New Zealand\\
\email{\{jeremy.rooney, keith.gordon\}@otago.ac.nz}\\
\and
Callaghan Innovation, Lower Hutt, New Zealand\\
\email{\{kirill.lagutin, andrew.mackenzie\}@callaghaninnovation.govt.nz}\\
\and
The New Zealand Institute for Plant and Food Research Limited, Nelson\\
\email{daniel.killeen@plantandfood.co.nz}
}

\maketitle              
\begin{abstract}


The rapid and accurate detection of biochemical compositions in fish is a crucial real-world task that facilitates optimal utilization and extraction of high-value products in the seafood industry. 
Raman spectroscopy provides a promising solution for quickly and non-destructively analyzing the biochemical composition of fish by associating Raman spectra with biochemical reference data using machine learning regression models.
This paper investigates different regression models to address this task and proposes a new design of Convolutional Neural Networks (CNNs) for jointly predicting water, protein, and lipids yield. 
To the best of our knowledge, we are the first to conduct a successful study employing CNNs to analyze the biochemical composition of fish based on a very small Raman spectroscopic dataset. 
Our approach combines a tailored CNN architecture with the comprehensive data preparation procedure, effectively mitigating the challenges posed by extreme data scarcity.
The results demonstrate that our CNN can significantly outperform two state-of-the-art CNN models and multiple traditional machine learning models, paving the way for accurate and automated analysis of fish biochemical composition.


\end{abstract}
\section{Introduction}


In New Zealand, Hoki and Mackerel are two commercially important deep-water fish species \cite{SpeciesSeafoodNZ}. Currently, they are primarily processed into low-value products such as fishmeal, limiting their economic potential.
The ability to extract high-value products, such as food-grade omega-3 fish oil and edible protein, depends heavily on understanding their biochemical compositions, particularly their water, protein and lipids yield contents. On average, water constitutes about 70-80\%, protein 10-20\%, and lipids yield 2-8\% of the fish's total weight \cite{coppespetricorenaChemicalCompositionFish2015}. 
However, these compositions can vary substantially, depending on different factors including catch location, season, and environmental conditions. 
This variability adds complexity to the quantification process, making it challenging to consistently determine the fish's biochemical makeup \cite{coppespetricorenaChemicalCompositionFish2015}.

Vibrational spectroscopic techniques like Fourier Transform Raman (FT-Raman), and InGaAs Raman at 1064 nm (InGaAs) spectroscopy provide rapid and non-destructive ways to analyze fish biochemical compositions~\cite{chengApplicationsNondestructiveSpectroscopic2013,heRecentAdvancesApplication2022,hassounQualityEvaluationFish2017}.
This capability makes it highly suitable for real-time high-throughput manufacturing processes \cite{heRecentAdvancesApplication2022}.
The spectral data obtained by Raman spectroscopy comprises of a series of features, each corresponding to the spectrum intensity at a given wavelength \cite{robinsonGeneticAlgorithmFeature2021}. The total number of wavelengths determines the total number of features. 
The spectral data provide a unique fingerprint-like pattern of peaks in the spectrum \cite{xuOverviewNondestructiveSpectroscopic2015}, enabling reliable prediction of fish biochemical compositions.


To effectively utilize the spectral data for fish biochemical composition analysis, this paper aims to develop an automated regression model that can accurately predict the biochemical composition of each catch in real-time. For this purpose, we acquire the biochemical reference data through traditional analytical chemistry methods. We also formulate a multi-output regression problem with the goal to simultaneously predict multiple biochemical components from the spectral features.

Previous studies have utilized traditional models such as Partial Least Squares Regression (PLSR), K Nearest Neighbor (KNN), Lasso Lars, Elastic Net, Support Vector Machines (SVM), Least Squares Support Vector Machines (LS-SVM), Light Gradient-Boosting Machine (LGBM), and Random Forest(RF), to address similar problems~\cite{heRecentAdvancesApplication2022,chengApplicationsNondestructiveSpectroscopic2013,hassounQualityEvaluationFish2017}.
However, the performance of these models is highly sensitive to preprocessing and feature selection operations, and they often fail to capture long-range dependencies among distance wavelengths, resulting in compromised accuracy  \cite{yangDeepLearningVibrational2019}. 

Recent work has shown that Convolutional Neural Networks (CNNs) can overcome these limitations by automatically preprocessing data, selecting features, and capturing complex patterns in spectral data\cite{mishraMultioutput1dimensionalConvolutional2022,yangDeepLearningVibrational2019,cuiModernPracticalConvolutional2018,passosTutorialAutomaticHyperparameter2022}. Upon using CNNs, normalization or standardization is typically utilized for data preprocessing. Meanwhile, performance can be enhanced further with additional preprocessing steps such as baseline correction, scatter correction and derivative transformation \cite{jernelvConvolutionalNeuralNetworks2020}. 

This study applies CNNs to Raman spectroscopy data to predict water, protein, and lipid yields contents in fish. Due to the specialized nature of this task, obtaining larger datasets is challenging and costly. To the best of our knowledge, no prior work has applied CNNs to such small Raman spectroscopic datasets. Our model is tailored to this domain and dataset size, addressing real-world constraints in fish composition analysis.
Key challenges include the limited dataset size, which restricts the model's ability to learn patterns, and the increased risk of overfitting \cite{poojaryEffectDataaugmentationFinetuned2021}. To address this, we develop a framework called FishCNN, which combines CNN with engineered regularization and data augmentation techniques to extract useful information and mitigate overfitting.
Our new developments leads to the following contributions.

\begin{enumerate}
    \item We develop an effective CNN model to jointly predict water, protein, and lipids yield contents from Raman spectroscopic data obtained from fish samples. 
    Our CNN model features a large kernel and small stride, which is unique and effective for our problem.


    \item We propose an effective procedure involving data preprocessing and data augmentation to train the CNN model effectively. Different from the common practice that applies data augmentation before preprocessing~\cite{bjerrumDataAugmentationSpectral2017,mishraMultioutput1dimensionalConvolutional2022}, 
    we apply data augmentation after preprocessing to ensure that the augmented data can reliably preserve important signal variations.

    \item We conduct comprehensive experiments to examine the performance of the proposed FishCNN
    , in comparison to traditional models and two state-of-the-art CNN models 
    \cite{mishraMultioutput1dimensionalConvolutional2022}. 
    Our experiments clearly show the performance advantage of our FishCNN in mitigating overfitting and improving prediction accuracy when dealing with small real-world datasets.
\end{enumerate}



\section{Related Work} \label{sec:background}


%


This section reviews relevant studies utilizing spectral data for the quantitative analysis of biochemical compositions in fish.
First, we examine studies focused specifically on fish biochemical analysis. 
Then, we broaden the scope to review general quantitative analysis approaches using spectral data.

Quantitative analysis of biochemical compositions in fish has been extensively studied using various spectroscopic techniques coupled with machine learning approaches \cite{heRecentAdvancesApplication2022,rohmanEmploymentAnalyticalTechniques2021,hassounQualityEvaluationFish2017,xuOverviewNondestructiveSpectroscopic2015}.
Traditional machine learning techniques like PLSR, KNN, LS-SVM, SVM and RF have been commonly employed to develop predictive models from spectral data for estimating various biochemical compositions in fish \cite{heRecentAdvancesApplication2022,chengApplicationsNondestructiveSpectroscopic2013}. However, these models often fail to capture long-range dependencies among distant wavelengths, resulting in compromised accuracy. 
Additionally, feature selection and data preprocessing are crucial but can introduce challenges such as the risk of removing informative features or overfitting due to collinearity \cite{yangDeepLearningVibrational2019}. Preprocessing corrects data artefacts and improves signal-to-noise ratios, but selecting the correct preprocessing pipeline remains complex due to variations in measurement environments and sample origins \cite{engelBreakingTrendsPreprocessing2013,DEBUS2021116459}.

Recent studies have increasingly explored deep learning, particularly one-dimensional convolutional neural networks (CNNs), for spectral data modeling~\cite{mishraMultioutput1dimensionalConvolutional2022,yangDeepLearningVibrational2019,cuiModernPracticalConvolutional2018,passosTutorialAutomaticHyperparameter2022}. 
Compared with traditional models, CNNs can automatically preprocess data and extract features, reducing the need for extensive preprocessing and feature selection, but these benefits are generally realized with large datasets~\cite{xuesongCommentaryReviewArticles2024}.
To address this challenge, a data augmentation method has been proposed as a solution to expand the sample space \cite{bjerrumDataAugmentationSpectral2017}, and its application in predicting drug contents from the Tablet NIR spectra dataset demonstrates its efficiency. 
Building on this, a recent study \cite{mishraMultioutput1dimensionalConvolutional2022} utilizes data augmentation and focuses on multi-output regression to simultaneously predict moisture content and soluble solids in pears using NIR spectra.
It adapts two benchmark CNN models (1CLNN and 3CLNN~\cite{cuiModernPracticalConvolutional2018,mishraMultioutput1dimensionalConvolutional2022}) with slight modifications to incorporate multi-output regression. 
However, we have noticed that these studies \cite{bjerrumDataAugmentationSpectral2017,mishraMultioutput1dimensionalConvolutional2022} only examine the CNN's performance by applying data augmentation before preprocessing. While this approach increases the training data size, it also risks amplifying data artefacts present in the raw spectral data. These artefacts can obscure underlying patterns and degrade model performance. 
To mitigate this issue, we propose applying data augmentation after preprocessing, ensuring that artefacts are minimized before augmentation, thus enhancing the quality of the training data and the robustness of the model.



In addition to typical CNNs, recent advances in spectral data analysis have explored transformer encoder-based architectures \cite{changRaTRamanTransformer2024} and hybrid models that combine one-dimensional convolution with multihead self-attention mechanisms for spectral data classification tasks \cite{renRamanConvMSANetHighAccuracy2023}. 
While these studies represent the current state-of-the-art, they focus on classification tasks with larger datasets. However, their techniques could inspire future work in quantitative analysis. 

\section{Proposed Approach} \label{Proposed Methods}
\vspace{-2.8em}
\begin{figure}
    \includegraphics[width=0.7\textwidth,center]{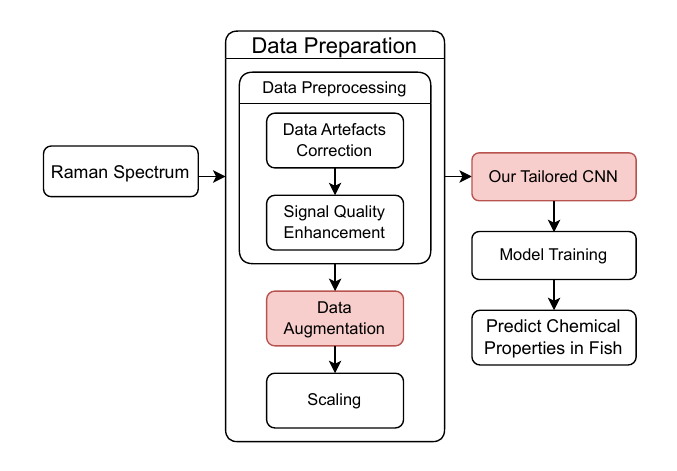}
    \caption{Overall Framework}
    \label{fig:overall-framework}
\end{figure}
\vspace{-1.5em}


We introduce a new framework, named FishCNN, which effectively integrates data preprocessing, data augmentation, and scaling techniques into a comprehensive procedure for training CNNs on small real-world spectroscopic datasets to predict water, protein, and lipids yield contents of fish samples.

The overall framework, illustrated in Fig. \ref{fig:overall-framework}, comprises several key components. 
First, the raw FT-Raman and InGaAs 1064 nm spectroscopy data ($X$) and the corresponding biochemical reference data ($Y$) are split into 6 folds for cross-validation. 

Next, the spectroscopy data is prepared using various methods, including data artefact correction, signal quality enhancement, extensive data augmentation, and scaling operations.
The specific data preparation steps are determined through a systematic trial-and-error evaluation process to identify the suitable combination that produces the highest mean \emph{cross-validated coefficient of determination} ($R^{2}CV$).


Subsequently, the prepared data are used to train our CNN model, enabling us to build the first CNN-based approach for predicting fish biochemical compositions. 
Through the synergy of data preprocessing and augmentation, the CNN model is expected to learn intricate patterns in Raman spectroscopic data and accurately predict the corresponding biochemical properties. 
For evaluation purposes, the cross-validated performance of FishCNN is compared to traditional models and two benchmark CNN models discussed in Section \ref{sec:background}. 


FishCNN introduces several innovative components designed to optimize the performance of CNNs on small spectral datasets, as highlighted in red in Fig. \ref{fig:overall-framework}. Specifically, we propose to use CNN with a large filter size and small stride, which can effectively extract useful features from small spectral datasets.
Different from common practice, we also decide to apply data augmentation after data preprocessing, ensuring that the augmented spectral data maintains high quality and relevance. This is empirically verified in Appendix \ref{appendix:framework components}. The specific details of each component are explained in the subsequent sections.

\subsection{Data Preprocessing} \label{sec:data preprocessing}

\vspace{-2.em}
\begin{figure}[h]
    \begin{subfigure}{.49\textwidth}
        \includegraphics[width=.9\linewidth]{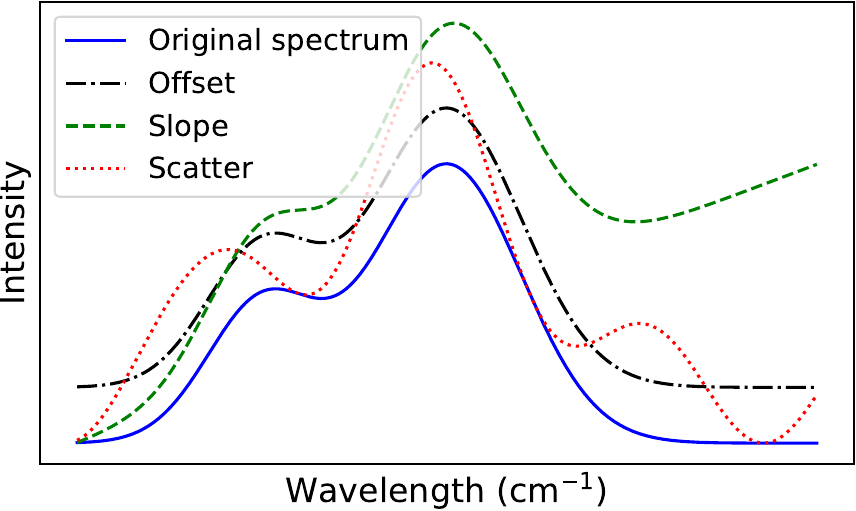} 
        \caption{Different data artefacts. The blue solid line represents the original spectrum. The red, black, and green dashed lines show scatter, baseline, and slope effects.}
        \label{fig:artefacts}
    \end{subfigure} 
    \hfill
    \hfill
    \begin{subfigure}{.49\textwidth}
        \includegraphics[width=.9\linewidth]{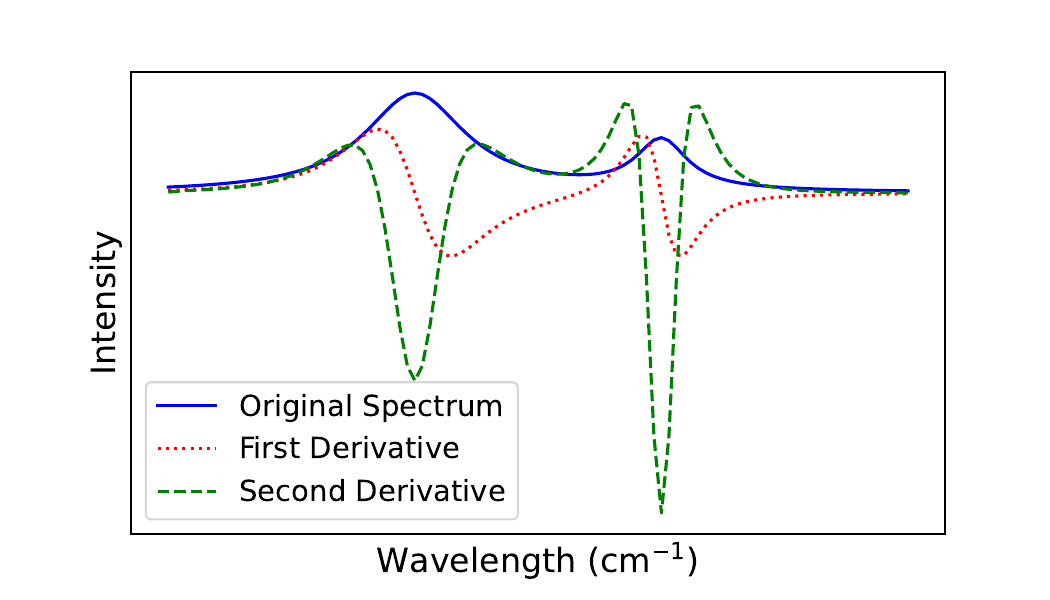}
        \caption{Resolving overlapping peaks with derivative transformation. The blue line is the original spectrum. The red and green lines are the first and second derivatives.}
        \label{fig:overlapping peaks}
    \end{subfigure}
    
    \caption{Example of data artefacts and derivatives in Raman spectroscopic dataset}
    \label{fig:artefacts and overlapping peaks}
\end{figure}



\vspace{-1.5em}
Data preprocessing is crucial for cleaning spectral signals and enabling the CNN model to effectively learn from limited samples. 
Raman spectroscopic data often contain data artefacts, like baseline shifts and light scatter, which can obscure important spectral signals \cite{engelBreakingTrendsPreprocessing2013}. Correctly handling these artefacts is essential for accurate analysis.
Furthermore, even if all artefacts are handled correctly, peaks from different biochemical components can overlap or interfere with each other, making it difficult to accurately identify and quantify individual compounds \cite{DEBUS2021116459,jernelvConvolutionalNeuralNetworks2020}. 

To address these challenges, a comprehensive data preprocessing procedure \cite{engelBreakingTrendsPreprocessing2013} is applied. 
As illustrated in Fig. \ref{fig:artefacts}, 
the blue line represents the artefact-free spectrum, while dashed lines represent variations. The Linear-Baseline (LB) \cite{engelBreakingTrendsPreprocessing2013} method removes baseline shifts (e.g. slopes and offsets), while the Standard Normal Variate (SNV) \cite{engelBreakingTrendsPreprocessing2013} corrects light scatter effects (i.e. scatters). Additionally, derivative transformation \cite{engelBreakingTrendsPreprocessing2013} enhances the spectral resolution quality and emphasizes relevant features by resolving overlapping peaks (see Fig. \ref{fig:overlapping peaks} for an example). It is applied after the data artefact correction since we want to enhance meaningful signals rather than amplify data artefacts. 


\vspace{-1.em}

\subsection{Selection of Data Preprocessing Methods}
\label{sec:preprocessing procedure selection}


The misuse of preprocessing methods can distort patterns within the spectrum signal \cite{yangDeepLearningVibrational2019}, thus, it is important to experiment with different combinations of preprocessing steps to identify the most effective procedure for any spectral dataset. 
For this purpose, we construct a preprocessing design matrix (see Appendix \ref{appendix:design matrix}), combining various signal artefact correction and signal enhancement methods. These methods include baseline correction, scatter correction, derivative transformations, and scaling. 
Each combination is applied in a predefined order based on~\cite{engelBreakingTrendsPreprocessing2013} to ensure consistency and effectiveness, avoiding the complexity associated with testing all possible permutations.


To determine the best preprocessing procedure, a trial-and-error approach is conducted, where the combination that enables CNNs to achieve the highest mean $R^2$ is selected.
The $R^2$ metric is chosen because it is a scale-invariant measure ranging from 0 to 1, indicating how well the model predicts the outcome variables. A recent study \cite{chiccoCoefficientDeterminationRsquared2021} has demonstrated that $R^2$ is more informative and does not suffer from the interpretability limitations of the MSE and RMSE metrics. This makes $R^2$ particularly suitable for tasks involving multiple outcome variables with different value ranges and distributions, as is the case in our study where we predict water, protein, and lipid contents simultaneously. The scale-invariant nature of $R^2$ allows for accurate comparison of model performance across these diverse outcome variables.

Following this systematic framework, the most suitable preprocessing procedure for any spectral dataset can be identified. This approach not only evaluates CNN performance but also validates and compares the effectiveness of preprocessing methods against the comparative models examined in Section \ref{sec:Comparison Methods}.





\subsection{Data Augmentation and Scaling} \label{sec:data augmentation and scaling}


Data augmentation techniques studied in \cite{bjerrumDataAugmentationSpectral2017} are applied to enhance the generalization ability and robustness of the trained CNN model.
The augmentation process introduces various irrelevant data artefacts to each spectrum instance, including baseline offsets, slopes, and multiplication differences illustrated in Fig. \ref{fig:artefacts}. By randomly applying these variations, the dataset is expanded, providing the CNN model with a more diverse set of instances to learn from.

We choose not to apply data augmentation to the raw unpreprocessed data because the data artefacts will also be amplified. This will result in large data variations, leading to the performance degradation of the CNN. The corresponding experiment results to support this argument can be found in Appendix \ref{appendix:framework components}. 
Therefore, applying data augmentation after pre-processing allows the model to learn from a more diverse dataset without introducing unnecessary noise or bias.

\begin{figure}
    \centering
    \includegraphics[width=.65\linewidth]{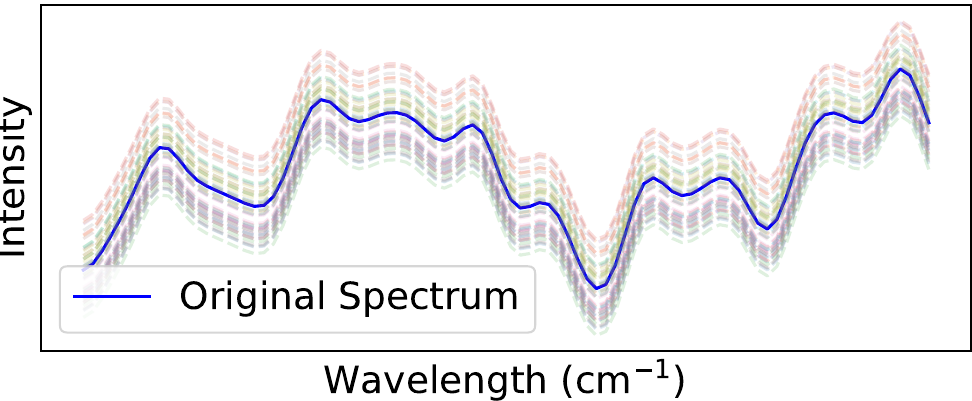}
    \caption{Example of data augmentation on a single spectrum instance. The original spectrum is shown as a solid blue line, while other dashed lines represent derived augmented spectra. }
    \label{fig:DA}
\end{figure}

To provide sufficient data diversity to effectively train the CNN, the training data is augmented 50 times. 
We choose an augmentation factor of 50 instead of 10 used in \cite{mishraMultioutput1dimensionalConvolutional2022,bjerrumDataAugmentationSpectral2017} because our dataset is extremely small. By increasing the augmentation factor to 50, a more diverse set of instances is generated, exposing the model to a wider range of variations during training. The diversity introduced by the augmented data, as illustrated in Fig. \ref{fig:DA}, helps the CNN model learn to differentiate between the original signal patterns and the introduced variations, enhancing its robustness and generalization capabilities (experiment results are reported in Appendix \ref{appendix:DA factor}
to support the choice of the augmentation factor ).

Before feeding into the CNN model, it is necessary to scale the data. This prevents essential low-intensity features from being overshadowed by less important high-intensity features due to scale differences \cite{yangDeepLearningVibrational2019,engelBreakingTrendsPreprocessing2013}. In spectral data, different wavelength regions can exhibit varying intensity levels or scales, with some regions (peaks) having much higher intensities than others, causing the model incorrectly prioritizes the high-intensity regions over low-intensity regions. To mitigate this issue, the global scaling method \cite{bjerrumDataAugmentationSpectral2017} is employed in our study to fit the training data, and then transform both the training and validation data for each cross-validation fold.

\vspace{-1.7em}

\vspace{-2.9em}

\subsection{Convolutional Neural Network Architecture (CNN)} \label{sec:CNN arch}



\begin{figure}
    \includegraphics[width=.8\textwidth,center]{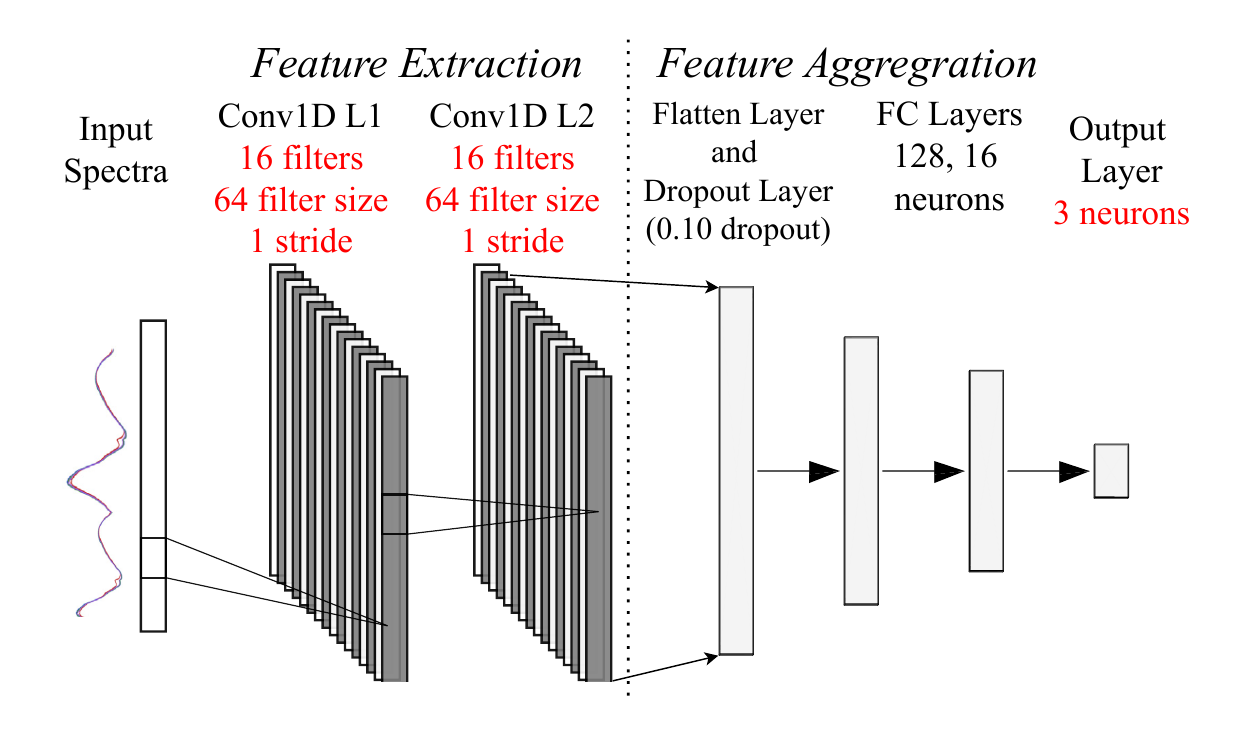}
    \caption{Proposed novel Large Kernel and Small Stride CNN Architecture for multi-output regression. It has two 1-dimensional convolutional (Conv1D) layers, one flatten layer, one dropout layer, two Fully Connected (FC) Layers, and one output layer. 
    Both two conv1D layers have the same special settings highlighted in red, with 16 filters, 64 large filter sizes, and 1 stride.
    }
    \label{fig:CNN architecture}
\end{figure}

Our 1-dimensional CNN architecture utilizes the augmented pre-processed data (see Section \ref{sec:preprocessing procedure selection}) as inputs, and performs multiple output regression to generate predicted target values including Water, Protein and Lipids yield contents as outputs. 
As illustrated in Fig. \ref{fig:CNN architecture}, our model adopts two stacked convolutional layers (labelled Conv1D L1 and Conv1D L2), a flattened layer with dropout (labelled as Flatten and Dropout Layer), two fully connected layers (labelled FC layers) and an output layer. 
To efficiently extract intricate patterns and aggregate them into high-level features to make predictions, the network can be divided into two blocks: \emph{Feature Extraction} and \emph{Feature Aggregration}. Both blocks are explained below.



\vspace{-2.em}

\subsubsection{Feature Extraction} \label{sec: Feature Extraction}



As highlighted in Fig. \ref{fig:CNN architecture}, our CNN model uniquely uses large filter sizes (64) with a small stride (1) in both convolutional layers (Conv1D L1, L2), with 16 filters and zero padding to maintain input dimensions. 
To the best of our knowledge, the use of this setting for one-dimensional CNN spectral analysis is novel since previous studies mainly consider small filter width and stride~\cite{mishraMultioutput1dimensionalConvolutional2022,cuiModernPracticalConvolutional2018}.
By stacking two convolutional layers with this setting, the network has a large receptive field size of 128, as each neuron in the final convolutional layer has a view of 128 consecutive data points. 
Compared to small filter widths and strides, our approach enables CNN to effectively capture intricate low-level patterns from the limited input sample space. 

\vspace{-.3em}

\subsubsection{Feature Aggregration} \label{sec:Feature Aggregration}

The flatten layer is used to concatenate output feature maps from the convolutional layers to create a one-dimensional embedding. Inspired by \cite{yangDeepLearningVibrational2019}, the random 10\% dropout layer is applied to the flatten output to mitigate the overfitting issue since the subsequent fully connected layers highly rely on feature maps extracted from convolutional layers. 

Subsequently, we have two fully connected layers and one output layer. The size of each FC layer is scaled down progressively from 128 to 16 to 3. This design is inspired by \cite{passosTutorialAutomaticHyperparameter2022} where the decreasing number of neurons creates a funneling effect, forcing the intricate low-level patterns learned in the convolutional layers to be aggregated to high-level abstractions. 

\newpage

\section{Design of the Experiments}

\subsection{Data Collection and Evaluation} \label{sec:data collection}
\vspace{-3.5em}
\begin{table}
    \centering
    \caption{Data Sets used in our study. Abbreviations: FT-Raman: Fourier Transform (FT) Raman; InGaAs-trun: InGaAs 1064 nm-truncated which excludes the glass vial range }
    \begin{tabularx}{.8\textwidth}{|>{\raggedright\arraybackslash}X|c|c|c|>{\raggedright\arraybackslash}X|} 
    \hline 
    Spectroscopy Data&  Samples&  Features & Wavelength Range ($\text{cm}^{-1}$) \\ \hline 
    FT-Raman& 39& 1971& 4001.81 - 202.533 \\ \hline 
    InGaAs-trun & 39& 427 &1891.58 - 580.109 \\ \hline
    \end{tabularx}
    \label{tab:datasets}
\end{table}
\vspace{-1.5em}

The detailed number of samples, spectral data ranges and the corresponding number of features are summarized in Table \ref{tab:datasets}. 
The Raman spectroscopic data used in this study are obtained using two different techniques: Fourier Transform (FT) Raman and InGaAs Raman at 1064 nm, both measuring the same set of fish samples.
While these two techniques share similar fundamentals, the InGaAs 1064 nm technique significantly reduces fluorescence background interference which is common in FT-Raman \cite{heHighFrequencyRamanAnalysis2020}. Since our fish samples are placed in glass vials, we excluded the wavelength range of 580.109 to 202.533 nm to avoid interference from the glass.

Although the InGaAs technique can provide a more accurate and reliable signal, as illustrated in Table \ref{tab:datasets}, it has fewer features, only covering the fingerprint region of the Raman spectrum, potentially limiting the information available for analysis \cite{heHighFrequencyRamanAnalysis2020}. Therefore, both spectral data are used in our experiments for comprehensive analysis, evaluating the performance of the CNN model on each to determine the most suitable dataset in this study.


Given the limited sample size, the dataset is split into 6 folds for cross-validation, with each fold containing approximately 32 samples for training and 7 samples for testing. As mentioned in Section \ref{sec:data augmentation and scaling}, we have applied 50 times data augmentation to the training data of each fold, while the testing data remains unchanged. 
To mitigate overfitting, we performed 10 individual runs of the model using 6-fold cross-validation. Additionally, we applied regularization techniques, including dropout (rate = 0.10) and L2 regularization to further enhance generalization on the small dataset.


\subsection{Comparison Methods} \label{sec:Comparison Methods}
\vspace{-.5em}

Several traditional models, including 
PLSR~\cite{heRecentAdvancesApplication2022}, 
KNN~\cite{heRecentAdvancesApplication2022}, 
Lasso Lars~\cite{pothExtensiveEvaluationMachine2022}, 
Elastic Net~\cite{pothExtensiveEvaluationMachine2022}, 
LS-SVM~\cite{heRecentAdvancesApplication2022}, 
Light Gradient Boosting Machine (LightGBM)~\cite{gaoMachineLearningPrediction2022}, Random Forest (RF)~\cite{gaoMachineLearningPrediction2022} are commonly used in the fields of spectroscopy. Therefore, we adopt them as baseline models in the experiments. 
For conciseness, only the best performing one will be presented, detailed descriptions of all baseline models are provided in Appendix~\ref{appendix:baseline performances}.


In addition, two state-of-the-art CNN models (1CLNN and 3CLNN 
\cite{cuiModernPracticalConvolutional2018,mishraMultioutput1dimensionalConvolutional2022}) for spectral analysis are included in the comparison. 
Both of these two CNN models have small filter sizes and strides, where 1CLNN has 1 convolutional layer, while 3CLNN has 3 convolutional layers. Their specific hyper parameter settings can be found in \cite{mishraMultioutput1dimensionalConvolutional2022}.
For a fair comparison, each model is trained and evaluated throughout our FishCNN framework based on the same dataset and experimental setup.
10 individual runs are performed and the pairwise Mann-Whitney U test \cite{robinsonGeneticAlgorithmFeature2021} with a significance level of 0.05 is used to test differences in $R^{2}CV$ between FishCNN and other models. 
Significant results are shown in bold in the tables.



\subsection{Parameter Settings for Our CNN Model}\label{sec:parameter settings}
\vspace{-3.3em}
\begin{table}[htbp]
    \centering
    \caption{Hyperparameter Settings for Our CNN Architectures}
    \begin{tabular}{|l|c|c|c|}
    \hline
    \textbf{Hyperparameter}           & \textbf{Our CNN}     \\
    \hline
    Convolutional Layer 1    & 16 filters, 64 filter size, 1 stride \\
    Convolutional Layer 2    & 16 filters, 64 filter size, 1 stride \\
    Fully Connected Layers   & [128, 16, 3]\\
    Dropout Rates            & 0.10         \\
    L2 Regularization $\beta$& 0.001        \\
    Batch Size               & 38           \\
    Initial Learning Rate    & 0.0015       \\
    Optimizer                & AdamW        \\
    Max Epochs              & 1500          \\
    Loss Function          & Huber   \\
    \hline
    \end{tabular}
    \label{tab:hyperparameters}
\end{table}
\vspace{-1.3em}

The parameter settings for our CNN model are summarized in Table \ref{tab:hyperparameters}. The batch size is set to 38, and the initial learning rate is 0.0015 according to the heustric formula: $learningrate = 0.01 * batch size / 256$  \cite{cuiModernPracticalConvolutional2018}.
We employ the AdamW optimizer, a variant of Adam with weight decay, which is often considered the best for spectral analysis \cite{yangDeepLearningVibrational2019}. 
Additionally, we implement early stopping, saving the best model when the validation loss does not decrease over 55 epochs \cite{passosTutorialAutomaticHyperparameter2022}.


\section{Results and Discussions} \label{sec:results}

In this section, we present the results of our FishCNN compared to all methods mentioned in Section \ref{sec:Comparison Methods}. 
While $R^2$ is the primary evaluation metric discussed here, additional detailed root mean squared error (RMSE) results are provided in Appendix~\ref{appendix:RMSE} for further analysis.
The results are divided into two parts. 
First, the overall performance averaging the results across all three targets (Water, Protein, Lipids yield) is presented in Section \ref{sec:Overall Results} to provide a general overview of the model's effectiveness.
Subsequently, the individual target performance, which examines each target separately, will be discussed in Section \ref{sec:individual target performance}.

To understand the multioutput $R^2$ performance, we use the formula: $R^2$: \[ R^2_{\text{overall}} = \frac{1}{N} \sum_{i=1}^{N} R^2_i \], where $R^2_{\text{overall}}$ is the overall $R^2$ value, $N$ is the number of targets, and $R^2_i$ is the $R^2$ value for each target $i$ (Water, Protein, Lipids yield). 
We calculate the $R^2_{\text{overall}}$ scores for each of the 6 folds in the cross-validation process. The reported $R^2_{\text{CV}}$ is expressed as the mean $\pm$ standard deviation of these 6 $R^2_{\text{overall}}$ scores. This allows us to evaluate the multioutput predictive performance while accounting for variance across the cross-validation folds.

\subsection{Overall Performance Analysis} \label{sec:Overall Results}
\vspace{-1em}
\begin{longtable}{|p{2cm}|p{2.5cm}|p{3cm}|p{1.2cm}|}
    \caption{Overall performance comparison of baseline, two benchmark CNNs, and our CNN. Both FT-Raman and InGaAs-trun are SNV \cite{engelBreakingTrendsPreprocessing2013} preprocessed. The p-values are obtained using the pairwise Mann-Whitney U test based on 10 individual runs, while deterministic models and FishCNN itself are denoted by "-". The significantly better results are highlighted in bold,  values before and after $\pm$ indicate mean$\pm$standard deviation.}
    \label{tab:overall performance}\\
    \hline
     Data          & Model    & $R^{2}CV$ (mean±std)   & \textit{p-value} \\
    \hline
    \endhead
     \multirow{4}{*}{FT-Raman}  & \textbf{FishCNN}       & \textbf{0.729±0.173} & -\\
                                 & 1CLNN \cite{cuiModernPracticalConvolutional2018} & 0.543±0.364    &    3.3e-04  \\
                                 & 3CLNN \cite{mishraMultioutput1dimensionalConvolutional2022} & 0.692±0.109	  &0.017        \\
                                 & KNN                 & 0.680±0.125	      & -     \\
\hline
    \multirow{4}{*}{InGaAs-trun}  & \textbf{FishCNN}                                                      & \textbf{0.811±0.105}  & -         \\
                                   & 1CLNN \cite{cuiModernPracticalConvolutional2018} & 0.769±0.073	 &    6.4e-05      \\
                                   & 3CLNN \cite{mishraMultioutput1dimensionalConvolutional2022} & 0.740±0.087  & 1.4e-08      \\
                                   & LS-SVM(poly)                                                & 0.752±0.091   & -        \\
    \hline

\end{longtable}

As shown in Table \ref{tab:overall performance}, on both FT-Raman and InGaAs truncated datasets, our CNN model achieves significantly better $R^{2}CV$ scores than all traditional models as well as 1CLNN and 3CLNN. The p-values of the relative pairwise Mann-Whitney U test \cite{robinsonGeneticAlgorithmFeature2021} are consistently lower than 0.05, indicating statistically significant improvement.

We notice that, although the InGaAs 1064 nm-truncated data has a smaller spectral range and fewer features (427) compared to the FT-Raman data (1971), our CNN model still outperforms the 1CLNN and 3CLNN models on both datasets. Additionally, the InGaAs-truncated data show a higher $R^{2}CV$ score than the FT-Raman data.
These results demonstrate the effectiveness of our CNN model in handling spectral data, even with extremely limited sample size.
To further investigate the robustness of our model, we also examine the individual target performance in Section \ref{sec:individual target performance} to identify the consistency of the model's performance across different targets.

\newpage
\subsection{Individual Target Performance} \label{sec:individual target performance}


\vspace{-1.2em}
\begin{small}
\begin{longtable}{|p{1.8cm}|p{2.5cm}|p{2.3cm}|p{2.75cm}|p{1.2cm}|}
    \caption{Each individual target performance comparison. For lipids yield, the InGaAs data is preprocessed by SNV followed by second order derivative with a 19-point window size, while others are all preprocessed by SNV only.  The p-values are obtained using the pairwise Mann-Whitney U test based on 10 individual runs, while deterministic models and FishCNN itself are denoted by "-". The significantly better results are highlighted in bold, values before and after $\pm$ indicate mean$\pm$standard deviation.}
    \label{tab:individual performance}\\
    \hline
     Target   & Data    & Model                    & $R^{2}CV$(mean±std) & \textit{p-value}  \\
    \hline
    \endhead
    \multirow{8}{*}{Water}    & \multirow{4}{*}{FT-Raman} & FishCNN       & 0.794±0.271   & -      \\
                               &                           & 1CLNN\cite{cuiModernPracticalConvolutional2018}      & 0.803±0.291   & 0.35      \\
                               &                           & 3CLNN\cite{mishraMultioutput1dimensionalConvolutional2022}  & 0.741±0.163    &  1.5e-05	    \\
                               
                               &                           & \textbf{RF}          & \textbf{0.832±0.080}  & 0.0018      \\

\cline{2-5}
                              & \multirow{4}{*}{InGaAs-trun} & \textbf{FishCNN}       & \textbf{0.919±0.035}   & -      \\
                               &                              & 1CLNN\cite{cuiModernPracticalConvolutional2018}      & 0.887±0.057      & 3.5e-04   \\
                               &                              & 3CLNN\cite{mishraMultioutput1dimensionalConvolutional2022} & 0.822±0.087	   &  1.1e-11    \\
                               &                              & LS-SVM(poly)                      & 0.824±0.058   & -      \\
\hline

    \multirow{8}{*}{Protein}  & \multirow{4}{*}{FT-Raman} & FishCNN       & 0.753±0.182 & -           \\
                               &                           & 1CLNN\cite{cuiModernPracticalConvolutional2018}   &  0.786±0.171	  & 0.8	           \\
                               &                           & 3CLNN\cite{mishraMultioutput1dimensionalConvolutional2022} & 0.698±0.153 & 3.6e-03          \\
                               &                           & KNN                  & 0.764±0.112     & -     \\
\cline{2-5}

                               & \multirow{4}{*}{InGaAs-trun}  & \textbf{FishCNN}   & \textbf{0.868±0.071} & -          \\
                               &                               & 1CLNN\cite{cuiModernPracticalConvolutional2018}       & 0.808±0.126		       & 4.4e-03    \\
                               &                               & 3CLNN\cite{mishraMultioutput1dimensionalConvolutional2022} & 0.742±0.130	    & 5.4e-09       \\
                               &                 & KNN                        & 0.798±0.065 & -           \\
\hline
    \multirow{8}{*}{Lipids yield} & \multirow{4}{*}{FT-Raman} & FishCNN        & 0.641±0.353   & -             \\
                             &                           & 3CLNN        & 0.637±0.187          & 1.0e-10      \\
                             &                           & 1CLNN       & 0.042±0.752           & 9.1e-09     \\
                             &                           & LS-SVM(linear)      & 0.643±0.331    & -            \\
\cline{2-5}
                            & \multirow{4}{*}{InGaAs-trun}  & \textbf{FishCNN}   & \textbf{0.847±0.104} & -                \\
                             &                               & 1CLNN\cite{cuiModernPracticalConvolutional2018}   & 0.693±0.196	   & 1.2e-06  \\
                             &                               & 3CLNN\cite{mishraMultioutput1dimensionalConvolutional2022} & 0.692±0.166	   & 2.0e-09             \\
                             &                               & SVR(linear)           & 0.722±0.215 & -                \\

    \hline

\end{longtable}
\end{small}




As illustrated in Table \ref{tab:individual performance}, the performance of each target obtained on the InGaAs dataset is always significantly better than those obtained on the FT-Raman dataset. This finding suggests that the InGaAs dataset is better than FT-Raman for predicting all three targets. This is likely due to the significantly reduced fluorescent background interference in InGaAs, despite of a small wavelength range~\cite{heHighFrequencyRamanAnalysis2020}.

Our FishCNN consistently outperforms all competing models across all three targets on the InGaAs dataset, demonstrating its effectiveness in handling spectral data and its generalization ability. 
However, FishCNN does not significantly outperform baseline models on the FT-Raman dataset for certain individual targets. This variation suggests that certain traditional models may be better suited for specific tasks or datasets with a wider spectral range and more features. Nevertheless, FishCNN consistently delivers strong results across all targets and datasets, reducing the need to determine suitable traditional models. This underscores the robustness and adaptability of FishCNN, even with a limited dataset size.
To illustrate the impact of kernel size of our FishCNN model, the corresponding ablation study can be found in Appendix \ref{sec:Ablation Study}.

Our results indicate that predicting lipids yield is more challenging compared to water and protein, as evidenced by its lower $R^{2}CV$ scores across all models and the necessity to apply an extra second-order derivative transformation with window size 19 after SNV correction. As illustrated in Section \ref{sec:data preprocessing}, SNV corrects data artefacts and second-order derivative further enhance the signal resolution quality.
This shows that advanced preprocessing techniques are crucial for correcting data artefacts and enhancing signal resolution in small Raman spectroscopic data, enabling accurate predictions of 
fish biochemical compositions.


\vspace{-.5em}

\section{Conclusions and Future Work}




In this paper, we successfully addressed the challenge of accurately predicting water, protein, and lipids yield contents in Hoki-Mackerel fish from Raman spectroscopic data. We developed the FishCNN framework that combines new CNN models with data preparation procedures including multiple data preprocessing and augmentation methods.
This approach was sepcifically designed for Raman spectroscopy-based fish biochemical composition analysis, a task that presents unique challenges due to the small dataset size and domain-specific requirements. While it is specialized for this application, FishCNN can be adapted to other spectral data with similar characteristics, particularly when data is limited. 

Our research yielded several key findings and contributions.
We discovered that preprocessing methods like SNV, followed by data augmentation, are crucial for enabling the CNN to work efficiently on small spectral datasets. 
We mitigated the risk of overfitting by conducting 10 individual runs of 6-fold cross-validation and applying data augmentation and regularization techniques. These strategies were essential for improving the model's generalization, adddressing the challenges posed by small datasets, which are common challenge in this field.
Additionally, a CNN configuration with a large kernel size and small stride consistently achieved high performance across all targets, demonstrating robustness in predicting the water, protein, and lipids yield contents in Hoki-Mackerel fish. This highlights its potential for real-time biochemical composition analysis in the marine fish industry. 
Furthermore, our work bridges a significant gap by applying CNNs to small spectral quantitative analysis data, an area that has been largely unexplored.
Our work opens new avenues for the application of deep learning in the analysis of fish spectral data, where large datasets are unavailable.

For future work, we aim to explore more advanced deep learning architectures such as transformer encoders \cite{changRaTRamanTransformer2024} and the combination of 1D convolutions with multihead self-attention mechanisms \cite{renRamanConvMSANetHighAccuracy2023}, which have already demonstrated the effectiveness in spectral classification tasks. 
Additionally, incorporating attention mechanisms could enhance interpretability by highlighting specific spectral features or absorption peaks, aligning with the domain principles.



\vspace{-.5em}

\section{Data, Code and Appendix Availability}

Due to confidentiality agreements, the dataset and code used in this study cannot be made publicly available. However, researchers interested in discussing our methods or seeking further clarification are encouraged to contact us directly. 
An online appendix is available, containing additional details such as preprocessing design matrix, baseline comparisons, RMSE scores, and ablation study, which can be accessed at \cite{}.


%
%


\bibliographystyle{splncs04}
\bibliography{./bib/main.bib}


%

\appendix
\include{appendix/appendix}

\end{document}

%% file: appendix/appendix.tex
\section{Supplementary Information} \label{sec:Appendix}

This supplementary material provides additional information and results to complement the main manuscript. It includes detailed performance results in terms of RMSE and $R^2$ for each traditional baseline model, presented for both the FT-Raman and InGaAs-truncated datasets. Additionally, this document contains further details on the experimental setups, data preprocessing techniques, and ablation studies not covered in the main manuscript.

\section{Preprocessing Design Matrix} \label{appendix:design matrix}

Table \ref{tab:preprocessing combinations} illustrates all 64 preprocessing procedures conducted in our experiments. It contains one raw data that has not been subjected to any preprocessing  methods, and other procedures that have been subjected to baseline correction, scatter correction, derivative transformations, and scaling preprocessing methods. 
Each procedure is applied in a predefined order based on~\cite{engelBreakingTrendsPreprocessing2013} to ensure consistency and effectiveness, avoiding the complexity associated with testing all possible permutations.

\begin{small}
\begin{longtable}{|c|c|c|c|c|}
    \caption{
    The experimental design matrix of 63 preprocessing procedures and one raw data. The preprocessing procedure selection has been used for both FT-Raman and InGaAs datasets. We use "-" to denote this step is not applied. 
    \newline Abbreviations: LB: Linear Baseline \cite{engelBreakingTrendsPreprocessing2013};
    GS: Global Scaling \cite{bjerrumDataAugmentationSpectral2017};
    SNV: Standard Normal Variate \cite{engelBreakingTrendsPreprocessing2013}; 
    }
    \label{tab:preprocessing combinations}    \\
\hline
    ID & Baseline  & Scatter  & Derivative (D) & Scaling (Sg) \\
\hline
\endhead
    1(raw) & -   & -   & -              & - \\
\hline
    2  & LB  & -   & -              & - \\
\hline
    3  & LB  & SNV & -              & - \\
\hline
    4  & LB  & SNV & 1st, w=5       & - \\
\hline
    5  & LB  & SNV & 1st, w=9       & - \\
\hline
    6  & LB  & SNV & 1st, w=13      & - \\
\hline
    7  & LB  & SNV & 1st, w=17      & - \\
\hline
    8  & LB  & SNV & 1st, w=21      & - \\
\hline
    9  & LB  & SNV & 1st, w=25      & - \\
\hline
    10 & LB  & SNV & 2nd, w=13      & - \\
\hline
    11 & LB  & SNV & 2nd, w=15      & - \\
\hline
    12 & LB  & SNV & 2nd, w=17      & - \\
\hline
    13 & LB  & SNV & 2nd, w=19      & - \\
\hline
    14 & LB  & SNV & 2nd, w=21      & - \\
\hline
    15 & LB  & SNV & 2nd, w=23      & - \\
\hline
    16 & LB  & SNV & 2nd, w=25      & - \\
\hline
    17 & LB  & SNV & 2nd, w=31      & - \\
\hline
    18 & -   & SNV & -              & - \\
\hline
    19 & -   & SNV & 1st, w=5       & - \\
\hline
    20 & -   & SNV & 1st, w=9       & - \\
\hline
    21 & -   & SNV & 1st, w=13      & - \\
\hline
    22 & -   & SNV & 1st, w=17      & - \\
\hline
    23 & -   & SNV & 1st, w=21      & - \\
\hline
    24 & -   & SNV & 1st, w=25      & - \\
\hline
    25 & -   & SNV & 2nd, w=13      & - \\
\hline
    26 & -   & SNV & 2nd, w=15      & - \\
\hline
    27 & -   & SNV & 2nd, w=17      & - \\
\hline
    28 & -   & SNV & 2nd, w=19      & - \\
\hline
    29 & -   & SNV & 2nd, w=21      & - \\
\hline
    30 & -   & SNV & 2nd, w=23      & - \\
\hline
    31 & -   & SNV & 2nd, w=25      & - \\
\hline
    32 & -   & SNV & 2nd, w=31      & - \\
\hline
    33 & LB  & -   & -              & GS \\
\hline
    34 & LB  & SNV & -              & GS \\
\hline
    35 & LB  & SNV & 1st, w=5       & GS \\
\hline
    36 & LB  & SNV & 1st, w=9       & GS \\
\hline
    37 & LB  & SNV & 1st, w=13      & GS \\
\hline
    38 & LB  & SNV & 1st, w=17      & GS \\
\hline
    39 & LB  & SNV & 1st, w=21      & GS \\
\hline
    40 & LB  & SNV & 1st, w=25      & GS \\
\hline
    41 & LB  & SNV & 2nd, w=13      & GS \\
\hline
    42 & LB  & SNV & 2nd, w=15      & GS \\
\hline
    43 & LB  & SNV & 2nd, w=17      & GS \\
\hline
    44 & LB  & SNV & 2nd, w=19      & GS \\
\hline
    45 & LB  & SNV & 2nd, w=21      & GS \\
\hline
    46 & LB  & SNV & 2nd, w=23      & GS \\
\hline
    47 & LB  & SNV & 2nd, w=25      & GS \\
\hline
    48 & LB  & SNV & 2nd, w=31      & GS \\
\hline
    49 & -   & SNV & -              & GS \\
\hline
    50 & -   & SNV & 1st, w=5       & GS \\
\hline
    51 & -   & SNV & 1st, w=9       & GS \\
\hline
    52 & -   & SNV & 1st, w=13      & GS \\
\hline
    53 & -   & SNV & 1st, w=17      & GS \\
\hline
    54 & -   & SNV & 1st, w=21      & GS \\
\hline
    55 & -   & SNV & 1st, w=25      & GS \\
\hline
    56 & -   & SNV & 2nd, w=13      & GS \\
\hline
    57 & -   & SNV & 2nd, w=15      & GS \\
\hline
    58 & -   & SNV & 2nd, w=17      & GS \\
\hline
    59 & -   & SNV & 2nd, w=19      & GS \\
\hline
    60 & -   & SNV & 2nd, w=21      & GS \\
\hline
    61 & -   & SNV & 2nd, w=23      & GS \\
\hline
    62 & -   & SNV & 2nd, w=25      & GS \\
\hline
    63 & -   & SNV & 2nd, w=31      & GS \\
\hline
    64 & -   & -   & -              & GS \\
\hline
\end{longtable}
\end{small}

\newpage
\section{RMSE results} \label{appendix:RMSE}

For this part, we present the same format as the main paper content, but for the RMSE metric. In particular, the results are divided into two parts. 
First, the overall RMSE performance averaging the results across all three targets (Water, Protein, Lipids yield) to provide a general overview of the model's effectiveness.
Subsequently, the individual target performance, which examines each target separately, will be discussed to provide a more detailed analysis of the model's performance on each target.

To understand the multioutput $RMSE$ performance, we use the formula: $RMSE$: \[ RMSE_{\text{overall}} = \frac{1}{N} \sum_{i=1}^{N} RMSE_i \], where $RMSE_{\text{overall}}$ is the overall $RMSE$ value, $N$ is the number of targets, and $RMSE_i$ is the $RMSE$ value for each target $i$ (Water, Protein, Lipids yield). 
We calculate the $RMSE_{\text{overall}}$ scores for each of the 6 folds in the cross-validation process. The reported $RMSE_{\text{CV}}$ is expressed as the mean $\pm$ standard deviation of these 6 $RMSE_{\text{overall}}$ scores. This allows us to evaluate the multioutput predictive performance while accounting for variance across the cross-validation folds.

\subsection{Overall Performance Analysis}

\begin{longtable}{|p{2cm}|p{2.5cm}|p{3.55cm}|p{1.2cm}|}
    \caption{Overall RMSE performance comparison of baseline, two benchmark CNNs, and our CNN. Both FT-Raman and InGaAs-trun are SNV \cite{engelBreakingTrendsPreprocessing2013} preprocessed. The p-values are obtained using the pairwise Mann-Whitney U test based on 10 individual runs, while deterministic models and FishCNN itself are denoted by "-". The significantly better results are highlighted in bold,  values before and after $\pm$ indicate mean$\pm$standard deviation.}
    \label{tab:overall RMSE performance}\\
\hline
  Data      & Model   & $RMSECV$(mean±std)   & \textit{p-value} \\
\hline
\endhead
\multirow{4}{*}{FT-Raman}   & \textbf{FishCNN} & \textbf{0.788±0.230}    &     -        \\
                            & 1CLNN \cite{cuiModernPracticalConvolutional2018}   & 0.900±0.340	    &     5.4e-02        \\
                            & 3CLNN \cite{mishraMultioutput1dimensionalConvolutional2022}   & 0.961±0.201	    &     5.4e-06        \\
                            & KNN    & 0.901±0.286    &     -        \\
\hline

    \multirow{4}{*}{InGaAs-trun}  & \textbf{FishCNN}                                                            & \textbf{0.650±0.120}         &   -     \\
                                  & 1CLNN \cite{cuiModernPracticalConvolutional2018}                 & 0.750±0.174	         &   7.3e-05     \\
                                  & 3CLNN \cite{mishraMultioutput1dimensionalConvolutional2022}        & 0.870±0.200         &   2.9e-10     \\
                                  & LS-SVM(poly) & 0.840±0.183         &   -     \\
\hline
\end{longtable}

As shown in Table \ref{tab:overall RMSE performance}, on both FT-Raman and InGaAs truncated datasets, our CNN model achieves significantly better $RMSECV$ scores than all traditional models as well as 1CLNN and 3CLNN.
The p-values of the relative pairwise Mann-Whitney U test \cite{robinsonGeneticAlgorithmFeature2021} are consistently lower than 0.05, indicating statistically significant improvement.  
This findings aligns with the results of the $R^{2}CV$ metric presented in the main text.


\subsection{Individual Target Performance}

\begin{small}

    \begin{longtable}{|p{1.8cm}|p{2.5cm}|p{2.3cm}|p{3.35cm}|p{1.2cm}|}
        \caption{Each individual target performance comparison. For lipids yield, the InGaAs data is preprocessed by SNV followed by second order derivative with a 19-point window size, while others are all preprocessed by SNV only.  The p-values are obtained using the pairwise Mann-Whitney U test based on 10 individual runs, while deterministic models and FishCNN itself are denoted by "-". The significantly better results are highlighted in bold, values before and after $\pm$ indicate mean$\pm$standard deviation.}
        \label{tab:individual RMSE performance}\\
        \hline
         Target   & Data    & Model                    & $RMSECV$(mean±std) & \textit{p-value}  \\
        \hline
        \endhead
        \multirow{8}{*}{Water}    & \multirow{4}{*}{FT-Raman} & \textbf{FishCNN}       & 1.104±0.520   & -      \\
                                   &                           & 1CLNN\cite{cuiModernPracticalConvolutional2018}      & 1.150±0.612	   &  0.46     \\
                                   &                           & 3CLNN\cite{mishraMultioutput1dimensionalConvolutional2022}  & 1.428±0.404	    &  3.9e-06    \\
                                   &                           & RF          & 1.216±0.413  & 0.0065   \\
    
    \cline{2-5}
                                  & \multirow{4}{*}{InGaAs-trun} & \textbf{FishCNN}       & \textbf{0.847±0.246}   & -      \\
                                   &                              & 1CLNN\cite{cuiModernPracticalConvolutional2018}      &  0.999±0.340	      & 1.6e-03  \\
                                   &                              & 3CLNN\cite{mishraMultioutput1dimensionalConvolutional2022} & 1.256±0.390	   &  1.7e-09      \\
                                   &                              & LS-SVM(poly)                      & 1.247±0.215   & -      \\
    \hline
    
        \multirow{8}{*}{Protein}  & \multirow{4}{*}{FT-Raman} & FishCNN       & 0.842±0.257 & -           \\
                                   &                           & 1CLNN\cite{cuiModernPracticalConvolutional2018}   &  0.794±0.247  &     8.4e-01           \\
                                   &                           & 3CLNN\cite{mishraMultioutput1dimensionalConvolutional2022} & 0.967±0.258	 & 3.0e-03        \\
                                   &                           & KNN                  & 0.879±0.287     & -     \\
    \cline{2-5}
    
                                   & \multirow{4}{*}{InGaAs-trun}  & \textbf{FishCNN}                                         & \textbf{0.648±0.215} & -          \\
                                   &                               & 1CLNN\cite{cuiModernPracticalConvolutional2018}           & 0.757±0.255		    &  5.3e-03 \\
                                   &                               & 3CLNN\cite{mishraMultioutput1dimensionalConvolutional2022} & 0.895±0.258	       &   1.3e-07      \\
                                   &                               & KNN                                                        & 0.817±0.200       & -           \\
    \hline
        \multirow{8}{*}{Lipids yield} & \multirow{4}{*}{FT-Raman} & \textbf{FishCNN}& \textbf{0.418±0.183}   & -             \\
                                 &                           & 3CLNN        & 0.757±0.326		          &  1.5e-10      \\
                                 &                           & 1CLNN       & -0.488±0.130           &   9.4e-03  \\
                                 &                           & LS-SVM(linear)      & 0.420±0.158    & -            \\
    \cline{2-5}
                                & \multirow{4}{*}{InGaAs-trun}  & \textbf{FishCNN}   & \textbf{0.304±0.090} & -                \\
                                 &                               & 1CLNN\cite{cuiModernPracticalConvolutional2018}   & 0.495±0.110		   & 7.4e-03  \\
                                 &                               & 3CLNN\cite{mishraMultioutput1dimensionalConvolutional2022} & 0.459±0.092	   &  2.9e-01            \\
                                 &                               & SVR(linear)           & 0.389±0.103 & -                \\
    
        \hline
    
\end{longtable}

\end{small}

As illustrated in Table \ref{tab:individual RMSE performance}, for RMSE performance, the performance of our FishCNN model is aligned with the $R^{2}CV$ results. This suggests that InGaAs-trun data is a better spectroscopic technique than FT-Raman in predicting Water, Protein, and Lipids yield. And our FishCNN model is more effective in predicting these targets than the benchmark models.

\newpage

\section{Detailed Baseline Models and Performance Results} \label{appendix:baseline performances}

In this appendix, we provide detailed performance results for all baseline models used in the study. While the main manuscript highlights the best-performing models, this appendix includes complete RMSE and $R^2$ results for all methods tested, allowing for a comprehensive evaluation.

\begin{small}

  \begin{longtable}{|p{1.8cm}|p{2.5cm}|p{2.3cm}|p{3.35cm}|p{3.2cm}|}
      
      \caption{RMSE and $R^2$ performance results for all baseline models. Both FT-Raman and InGaAs-trun are SNV \cite{engelBreakingTrendsPreprocessing2013} preprocessed. Values before and after $\pm$ indicate mean$\pm$standard deviation.}
      \label{tab:baseline models performance}\\
      \hline
       Target   & Data    & Model             & $R^{2}CV$(mean±std) & $RMSECV$(mean±std)   \\
      \hline
      \endhead
      \multirow{9}{*}{Overall}   & \multirow{4}{*}{FT-Raman}  & LS-SVM(poly)       & 0.594±0.197           & 1.023±0.198   \\  
                                                              && LS-SVM(linear)     & 0.657±0.145           & 0.967±0.158   \\  
                                                              && LS-SVM(rbf)         & 0.368±0.174           & 1.367±0.252   \\  
                                                              && PLSR1               & 0.569±0.174           & 1.024±0.148   \\  
                                                              && ENet                & 0.675±0.137           & 0.958±0.160   \\  
                                                              && LassoLars           & 0.642±0.124           & 1.043±0.172   \\  
                                                              && LGBM                & 0.633±0.132           & 0.952±0.146   \\  
                                                              && KNN                 & 0.680±0.125            & 0.901±0.286   \\  
                                                              && RF                  & 0.598±0.145           & 1.024±0.195   \\  
  
  \cline{2-5}
                                 & \multirow{4}{*}{InGaAs-trun} & LS-SVM(poly)                    & 0.752±0.091           & 0.840±0.183             \\
                                                                && LS-SVM(linear)                   & 0.711±0.078           & 0.957±0.094             \\
                                                                && LS-SVM(rbf)                   & 0.684±0.085           & 0.931±0.174             \\
                                                                && PLSR1                    & 0.732±0.137           & 0.885±0.145             \\
                                                                && ENet                     & 0.706±0.076           & 0.966±0.094             \\
                                                                && LassoLars                & 0.678±0.092           & 1.001±0.081             \\
                                                                && LGBM                     & 0.650±0.108           & 1.020±0.169             \\
                                                                && KNN                      & 0.711±0.067           & 0.864±0.209             \\
                                                                && RF                       & 0.624±0.112           & 1.008±0.191             \\
  \hline

      \multirow{9}{*}{Water}    & \multirow{4}{*}{FT-Raman} & LS-SVM(poly)    & 0.742±0.247         & 1.353±0.377       \\   
                                                            && LS-SVM(linear)  & 0.752±0.191         & 1.374±0.282       \\   
                                                            && LS-SVM(rbf)      & 0.528±0.196         & 2.020±0.333       \\   
                                                            && PLSR1            & 0.761±0.190         & 1.344±0.310       \\   
                                                            && ENet             & 0.749±0.196         & 1.379±0.293       \\   
                                                            && LassoLars        & 0.697±0.231         & 1.518±0.332       \\   
                                                            && LGBM             & 0.827±0.053         & 1.249±0.256       \\   
                                                            && KNN              & 0.785±0.070         & 1.393±0.309       \\   
                                                            && RF               & 0.832±0.080         & 1.216±0.413       \\   
  
  \cline{2-5}
                                & \multirow{4}{*}{InGaAs-trun}  & LS-SVM(poly)                    & 0.824±0.058        & 1.247±0.215             \\
                                                                && LS-SVM(linear)                   & 0.781±0.054         & 1.395±0.169             \\
                                                                && LS-SVM(rbf)                   & 0.799±0.072         & 1.337±0.367             \\
                                                                && PLSR1                    & 0.794±0.082         & 1.332±0.234             \\
                                                                && ENet                     & 0.781±0.054         & 1.397±0.170             \\
                                                                && LassoLars                & 0.766±0.038         & 1.453±0.142             \\
                                                                && LGBM                     & 0.725±0.115         & 1.548±0.372             \\
                                                                && KNN                      & 0.846±0.081         & 1.167±0.406             \\
                                                                && RF                       & 0.748±0.122         & 1.458±0.471             \\
  \hline
  
      \multirow{8}{*}{Protein}  & \multirow{4}{*}{FT-Raman} & LS-SVM(poly)       & 0.542±0.250           & 1.179±0.266       \\   
                                                            && LS-SVM(linear)     & 0.644±0.195           & 1.043±0.260       \\   
                                                            && LS-SVM(rbf)         & 0.490±0.184           & 1.282±0.275       \\   
                                                            && PLSR1               & 0.599±0.180           & 1.121±0.230       \\   
                                                            && ENet                & 0.644±0.196           & 1.043±0.259       \\   
                                                            && LassoLars           & 0.573±0.227           & 1.151±0.302       \\   
                                                            && LGBM                & 0.682±0.146           & 0.999±0.205       \\   
                                                            && KNN                 & 0.764±0.112            & 0.879±0.287       \\   
                                                            && RF                  & 0.664±0.173           & 1.022±0.251       \\   
  \cline{2-5}
  
                                 & \multirow{4}{*}{InGaAs-trun}  & LS-SVM(poly)                   & 0.708±0.132           & 0.942±0.188               \\  
                                                                && LS-SVM(linear)                   & 0.658±0.103           & 1.042±0.151               \\ 
                                                                && LS-SVM(rbf)                   & 0.768±0.085           & 0.875±0.226               \\ 
                                                                && PLSR1                    & 0.701±0.204           & 0.908±0.225               \\ 
                                                                && ENet                     & 0.647±0.106           & 1.059±0.156               \\ 
                                                                && LassoLars                & 0.641±0.097           & 1.073±0.151               \\ 
                                                                && LGBM                     & 0.711±0.091           & 0.965±0.175               \\ 
                                                                && KNN                      & 0.798±0.065           & 0.817±0.200               \\ 
                                                                && RF                       & 0.730±0.096           & 0.932±0.216               \\ 
  \hline
      \multirow{8}{*}{Lipids yield} & \multirow{4}{*}{FT-Raman} & LS-SVM(poly)            & 0.497±0.327                & 0.535±0.132                     \\
                                                                && LS-SVM(linear)          & 0.643±0.331                & 0.420±0.158                     \\
                                                                && LS-SVM(rbf)              & 0.087±0.259                & 0.800±0.178                     \\
                                                                && PLSR1                    & 0.346±0.431                & 0.608±0.123                     \\
                                                                && ENet                     & 0.632±0.274                & 0.451±0.114                     \\
                                                                && LassoLars                & 0.655±0.204                & 0.460±0.090                     \\
                                                                && LGBM                     & 0.391±0.345                & 0.609±0.146                     \\
                                                                && KNN                      & 0.403±0.193                & 0.640±0.123                     \\
                                                                && RF                       & 0.349±0.419                & 0.643±0.162                     \\
  \cline{2-5}
                              & \multirow{4}{*}{InGaAs-trun}  & LS-SVM(poly)                   & 0.727±0.200                 & 0.392±0.101                     \\
                                                              && LS-SVM(linear)                   & 0.693±0.180                & 0.434±0.081                     \\
                                                              && LS-SVM(rbf)                   & 0.483±0.252                & 0.581±0.145                     \\
                                                              && PLSR1                    & 0.702±0.211                & 0.414±0.111                     \\
                                                              && ENet                     & 0.690±0.168                & 0.441±0.075                     \\
                                                              && LassoLars                & 0.628±0.228                & 0.477±0.090                     \\
                                                              && LGBM                     & 0.515±0.248                & 0.548±0.163                     \\
                                                              && KNN                      & 0.494±0.165                & 0.594±0.144                     \\
                                                              && RF                       & 0.395±0.237                & 0.635±0.129                     \\
  
      \hline
  
\end{longtable}

\end{small}

\newpage

\section{Ablation Study} \label{sec:Ablation Study}

In this section, we present the ablation study results to evaluate the effectiveness of the individual components of our proposed framework, including the order of applying Data Augmentation (DA) and preprocessing, and the impact of the augmentation factor. 

We conduct the ablation study on the InGaAs-truncated dataset with the focus on the overall $R^{2}CV$ performance. InGaAs data is chosen for this study due to its better performance compared to FT-Raman data, as shown in the main text. While, the overall $R^{2}CV$ is used since the multioutput regression task is the main focus of this study.

\subsection{Effectiveness of the Framework Components}  \label{appendix:framework components}

\vspace{-1cm}

\begin{figure}
    \centering
    \includegraphics[width=.9\textwidth,left]{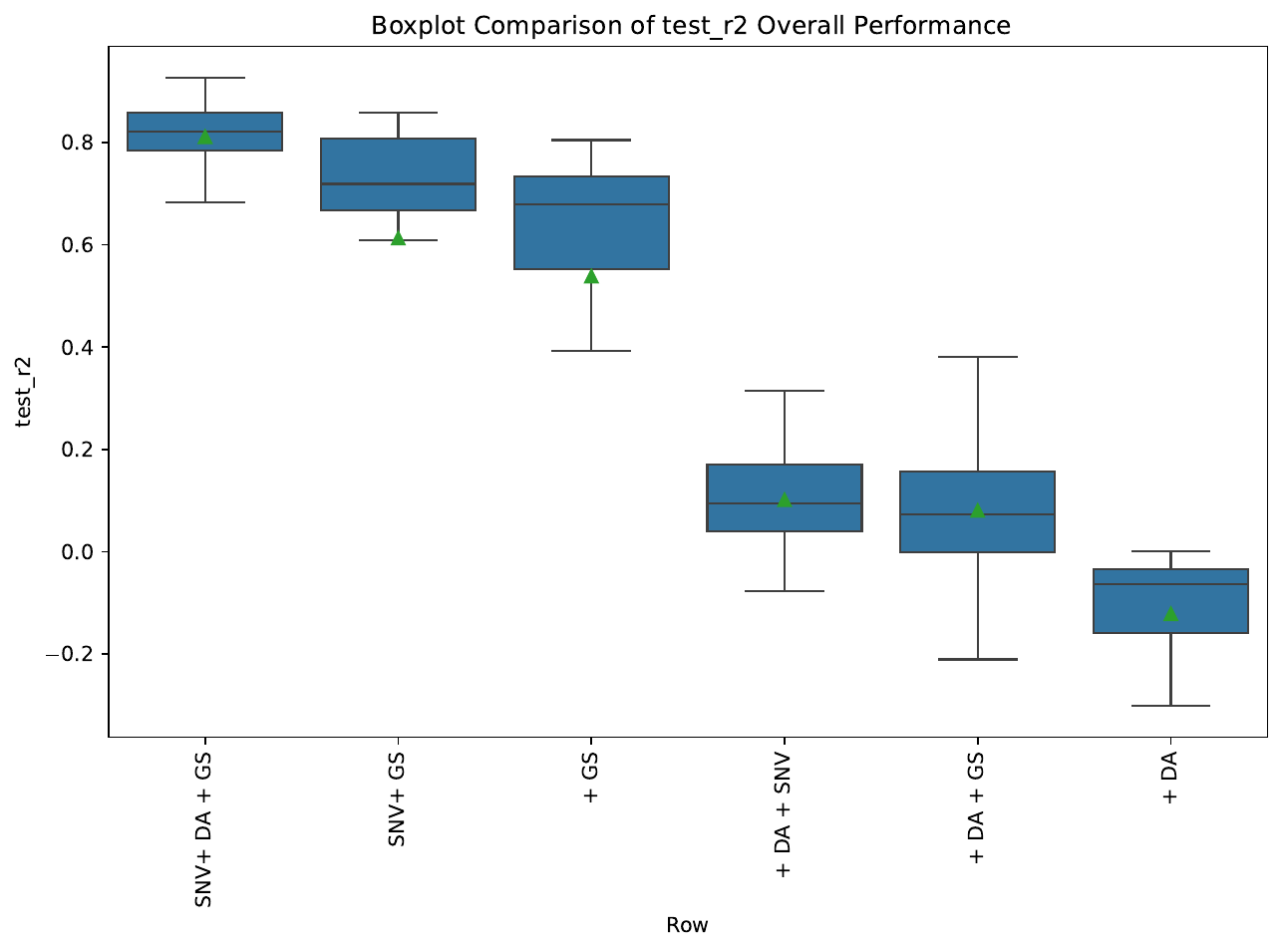}
    \caption{Boxplot of Effectiveness of Framework Components. The mean value is denoted as the green triangle, and the median is denoted as the line that split the box in two.  X-axis: Different DA Factors, Y-axis: $R^2$ Performance}
    \label{fig:Effectiveness of Framework Components}
\end{figure}

As illustrated in Table \ref{tab:framework component ablation study} and Fig.\ref{fig:Effectiveness of Framework Components}, we have found that the sequential application of preprocessing and data augmentation (SNV+DA+GS, SNV+DA) perform significantly better than others, especially when compared to the common practice \cite{bjerrumDataAugmentationSpectral2017,mishraMultioutput1dimensionalConvolutional2022} of applying data augmentation before preprocessing (DA+SNV). 
The p-values of the relative pairwise Mann-Whitney U test \cite{robinsonGeneticAlgorithmFeature2021} are consistently lower than 0.05, indicating statistically significant improvement. 
The reason behind this is that data augmentation introduces noise and irrelevant information to the raw unpreprocessed data, which further amplifies the noise and makes it difficult for the model to learn meaningful patterns. 

In addition to the common practise, we have investigated the impact of not applying the SNV preprocessing method and data augmentation (DA) on the model's performance. And the results clearly show a significant drop in performance when the SNV preprocessing method is excluded (DA+GS, DA) and when data augmentation is excluded (SNV+GS, GS).


This ablation study underscores the contribution of each component in preparing spectral data for predicting multiple fish biochemical contents effectively. The results demonstrate that the SNV preprocessing method and data augmentation must be applied sequentially to enhance the model's performance and generalization ability, especially when dealing with ultra-small spectral datasets.

\vspace{-.4cm}
\begin{table}
    \centering
    \caption{Ablation Study: Effectiveness of Framework Components in InGaAs-truncated Data. The p-values are obtained using the pairwise Mann-Whitney U test between SNV+DA+GS and others \newline Abbreviations: SNV: Standard Normal Variate preprocessing method, DA: Data Augmentation, GS: Global Scaling, +: sequential application}
    \label{tab:framework component ablation study}
    \begin{tabular}{|c|c|c|}
        \hline
        Procedure & $R^{2}CV$ & \textit{p-value} \\
        \hline
        SNV+DA+GS & 0.811±0.105	 & - \\
        SNV+GS & 0.613±0.415     & 1.6e-08 \\
        GS    & 0.539±0.372	  & 2.5e-16	 \\
        DA+SNV & 0.101±0.107  & 6.5e-21 \\
        DA+GS & 0.080±0.115	  & 5.9e-21	 \\
        DA    & -0.122±0.132  & 1.8e-21	 \\
        \hline
    \end{tabular}
\end{table}

\subsection{Effectiveness of Different Data Augmentation Factor} \label{appendix:DA factor}

\begin{figure}
    \centering
    \includegraphics[width=0.9\textwidth]{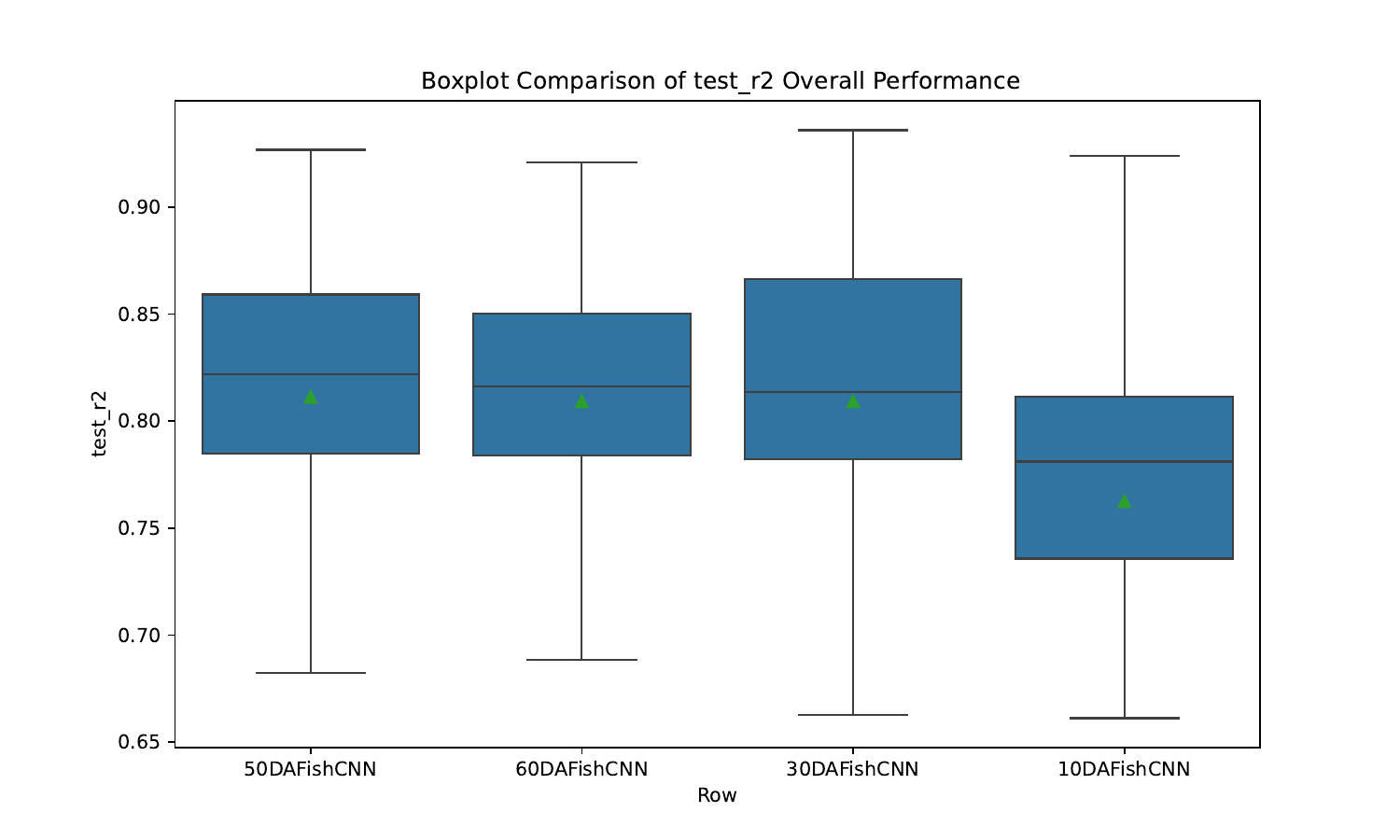}
    \caption{DA Factor Boxplot. The mean value is denoted as the green triangle, and the median is denoted as the line that split the box in two. \newline X: Different DA Factors, Y: $R^2$ Performance}
    \label{fig:effect da}
\end{figure}

The data augmentation factor determines the number of synthetic samples generated from each original sample, thereby increasing the diversity of the training data. A higher augmentation factor exposes the model to a wider range of variations during training, which can potentially enhance its robustness and generalization capabilities. To investigate the effect of the augmentation factor, we evaluate the performance of our framework using augmentation factors of 10, 30, 50, and 60 on the InGaAs-truncated dataset.

Table \ref{tab:DA factor ablation study} presents the results of this ablation study. As shown, the model's performance is significantly improved when the augmentation factor is increased from 10 to 50. The p-values obtained from the pairwise Mann-Whitney U test between the augmentation factor of 50 and other factors indicate that the improvement is statistically significant (p-value < 0.05) for the factor of 10, but not for factors of 30 and 60. 

To determine which factor should be chosen, the boxplot, as shown in Fig. \ref{fig:effect da}, is used to explore their performance distribution. We can clearly observe that the mean and median of an augmentation factor of 50 is slightly better than others. Hence, 50 is chosen as the optimal value for our framework.


\begin{table}
    \centering
    \caption{Ablation Study: Comparsions of 10, 30, 50, 60 Data Augmentation Factor in InGaAs-truncated Data. The p-values are obtained using the pairwise Mann-Whitney U test between 50 DA and others }
    \label{tab:DA factor ablation study}
    \begin{tabular}{|c|c|c|}
        \hline
        Procedure             & $R^{2}CV$              & \textit{p-value}  \\
        \hline
        SNV+ 50DA + GSX      & 0.811±0.105            &  -        \\
        SNV+ 60DA + GSX      & 0.809±0.085            &  0.25        \\
        SNV+ 30DA + GSX      & 0.809±0.092            &   0.38       \\
        SNV+ 10DA + GSX      & 0.763±0.163            &  0.00055 \\
        \hline
    \end{tabular}
\end{table}




\newpage

\subsection{Effect of Different Convolutional Kernel Size} \label{appendix:kernel size}


Based on the comparative analysis of various kernel sizes illustrated in Table \ref{tab:SNV kernel_sizes_comparison}, it is evident that the choice of kernel size significantly impacts performance metrics across different targets. According to the results, the 64 kernel size demonstrates consistently high performance across key metrics, particularly in water and protein prediction performance. However, the prediction accuracy for the lipids yield target with a 64 kernel size is not optimal. 
As discussed in Section \ref{sec:individual target performance}, predicting lipids yield is more challenging and requires additional derivative transformations. Therefore, we conducted further experiments using SNV + second derivative transformation with a 19-window size, as illustrated in Table \ref{tab:2nd d kernel_sizes_comparison}. 
These results show that the 64 kernel size consistently outperforms other sizes in this context, reaffirming its suitability as the optimal choice for balancing model complexity and predictive accuracy.

\begin{table}[ht]
\centering
\caption{Ablation Study: Comparsion Results of 64, 16, 8, 4 convolutional Kernel sizes on InGaAs-truncated data. The data preparation procedure is SNV + DA + GS. The p-values are obtained using the pairwise Mann-Whitney U test between 64 sizes and others }
\label{tab:SNV kernel_sizes_comparison}
\begin{tabular}{|r|l|l|l|l|l|}
    \hline
    Performance \& \textit{p-value} & 64Size                      & 16Size           & 8Size            & 4Size            \\
    \hline
    Overall $R^{2}CV$               & 0.811±0.098            & 0.786±0.149      & 0.808±0.080      & 0.809±0.066      \\
    \textit{p-value}                & -                          & 9.5e-02          & 1.4e-01          & 1.3e-01          \\
    \hline
    Water $R^{2}CV$                 & 0.917±0.038            & 0.910±0.041      & 0.905±0.039      & 0.897±0.056      \\
    Water \textit{p-value}          & -                         & 1.4e-01          & 8.8e-03          & 5.4e-03          \\
    \hline
    Protein $R^{2}CV$               & 0.867±0.070            & 0.802±0.279      & 0.846±0.089      & 0.840±0.097      \\
    Protein \textit{p-value}        & -                          & 3.1e-02          & 1.0e-01          & 5.6e-02          \\
    \hline
    Lipids yield $R^{2}CV$          & 0.649±0.291            & 0.647±0.257      & 0.675±0.236      & 0.690±0.176      \\
    Lipids yield \textit{p-value}   & -                          & 4.9e-01          & 7.1e-01          & 7.4e-01          \\
    \hline
\end{tabular}
\end{table}

\vspace{-1cm}

\begin{table}
\centering
\caption{Ablation Study: Comparsion Results of 64, 16, 8, 4 convolutional Kernel sizes on InGaAs-truncated data. The data preparation procedure is SNV + 2nd derivative with 19 window sizes + DA + GS. The p-values are obtained using the pairwise Mann-Whitney U test between 64 sizes and others }
\label{tab:2nd d kernel_sizes_comparison}
\begin{tabular}{|r|l|l|l|l|}
    \hline
    Hyperparameters         & 64Size            & 16Size           & 8Size            & 4Size            \\
    \hline
    Overall $R^{2}CV$       & 0.758±0.095       & 0.738±0.117      & 0.738±0.104      & -0.874±14.851    \\
    Overall \textit{p-value} & -                & 1.1e-01          & 1.0e-01          & 3.5e-02          \\
    \hline
    Water $R^{2}CV$         & 0.565±0.268       & 0.540±0.257      & 0.542±0.250      & -4.205±44.469    \\
    Water \textit{p-value}  & -                & 1.4e-01          & 1.4e-01          & 1.1e-01          \\
    \hline
    Protein $R^{2}CV$       & 0.862±0.101       & 0.845±0.165      & 0.842±0.151      & 0.807±0.261      \\
    Protein \textit{p-value} & -               & 5.6e-01          & 4.4e-01          & 4.6e-01          \\
    \hline
    Lipids yield $R^{2}CV$  & 0.846±0.102       & 0.828±0.151      & 0.831±0.119      & 0.777±0.231      \\
    Lipids yield \textit{p-value} & -          & 2.5e-01          & 2.5e-01          & 6.0e-03          \\
    \hline
\end{tabular}
\end{table}